\documentclass[runningheads,a4paper]{llncs}

\usepackage{epstopdf}
\usepackage{textcomp}
\usepackage{amssymb}
\usepackage{graphicx}
\usepackage{framed}
\usepackage{color}
\usepackage{changepage}

\usepackage{url}
\urldef{\mailsa}\path|{asahu}@rhrk.uni-kl.de|

\newcommand{\keywords}[1]{\par\addvspace\baselineskip
\noindent\keywordname\enspace\ignorespaces#1}

\begin{document}

\mainmatter  

\title{Survey of reasoning using Neural Networks}

\titlerunning{Survey of reasoning using Neural Networks}

%
%
\author{Amit Sahu}
\authorrunning{Survey of reasoning using Neural Networks}

\institute{TU Kaiserslautern, Informatik,\\
Erwin-Schr\"{o}dinger-Stra\ss{}e 1, 67663 Kaiserslautern, Germany\\
\mailsa\\
\url{http://www.informatik.uni-kl.de}}

%
%

\toctitle{Survey of reasoning using Neural Networks}
\tocauthor{Amit Sahu}
\maketitle

\begin{adjustwidth}{-5mm}{}
\begin{abstract}
Reason and inference require process as well as memory skills by humans. Neural networks are able to process tasks like image recognition (better than humans) but in memory aspects are still limited (by attention mechanism, size). Recurrent Neural Network (RNN) and it's modified version LSTM are able to solve small memory contexts, but as context becomes larger than a threshold, it is difficult to use them. The Solution is to use large external memory. Still, it poses many challenges like, how to train neural networks for discrete memory representation, how to describe long term dependencies in sequential data etc. Most prominent neural architectures for such tasks are Memory networks: inference components combined with long term memory and Neural Turing Machines: neural networks using external memory resources. Also, additional techniques like attention mechanism, end to end gradient descent on discrete memory representation are needed to support these solutions. Preliminary results of above neural architectures on simple algorithms (sorting, copying) and Question Answering (based on story, dialogs) application are comparable with the state of the art. In this paper, I explain these architectures (in general), the additional techniques used and the results of their application. 
 
\end{abstract}

\section{Introduction}
Artificial Intelligence has two grand challenges: build models that can make multiple computational steps in answering a question and models that can describe long term dependence in sequential data. Most machine learning models lack in the ability to easily read and write to a memory (large) component and infer using a small part of this large memory. For example, tasks to answer questions from a set of facts in a story, to watch a movie and answer questions about it or to conduct dialogues cannot be solved by these models. In principle they can be solved by a language modeler such as a recurrent neural network (RNN) (\cite{intRNN}; \cite{lstm}), but their memory is too small and not compartmentalized enough to remember the required facts accurately. Even in simple memorization task like copying a just seen sequence RNNs are known to have problems \cite{memoprob}. RNNs are turing complete \cite{TCRNN} and therefore have the capacity to simulate arbitrary procedures but in practice they are not able to.

In this survey, I discuss and compare some of the proposed solutions to these problems. In Neural Turing Machine (NTM) \cite{ntm}, these problems are attempted in analogy to Turing's enrichment of finite-state machine by an infinite memory tape. NTM work like a working memory by solving tasks that require the application of approximate rules to \textquotedblleft rapidly-created variables \textquotedblright \cite{rapidvar} and by using an attentional process to read from and write to memory selectively. In Memory Networks \cite{memnn}, the idea is to combine successful machine learning strategies with memory component that can be read and written to. End-to-end memory networks (MemN2N) \cite{etemn} extends on memory networks by removing problem in backpropagation and requirement of strong supervision at each layer. It is a continuous model that only require input-output pairs in comparison to memory network which require supporting facts in memory (only during training) as well.

\subsubsection*{Paper Structure}
The organization of this survey paper begins with a brief background about basic Neural architectures that use some kind of memory for reasoning and inference (section \ref{sec:background}). It follows with explanation of some of the prominent approaches in using large memory (section \ref{sec:approaches}). Some of the additional techniques like memory focus and their continuous representation are discussed along with the approaches. After approaches, the experiments done using them, their results and conclusions are discussed (section \ref{sec:experiments}). Finally, conclusion on comparison is drawn from experiments about these approaches (section \ref{sec:conclusion}).   

\section{Background} \label{sec:background}

\subsection{RNN}
Recurrent Neural Networks are neural networks with loop at a hidden node i.e. output of the hidden node is put back into the hidden node alongwith the input at next timestamp. Thus, the output of hidden node acts like a dynamic state whose evolution depends on both the input and current state (output of hidden node at previous timestamp). By unfolding RNN through time one can perceive that the context (dynamic state) from an earlier timestamp could affect the behaviour of the network at later timestamps. 
\\
RNN give way to \textquotedblleft vanishing and exploding gradient \textquotedblright problem. As the gradient moves across timestamps in backpropagation it's multiplied with $w_{lh}(t)$ (weight of loop link) depending on it's value it vanishes ($w_{lh}(t) < 1$) or explodes ($w_{lh}(t) > 1$). Thus, RNN can unfold into a limited number of timetamps, as increasing after won't have any effect because of this problem.
\subsection{LSTM}
\begin{figure}
\centering
\includegraphics[height=6.2cm]{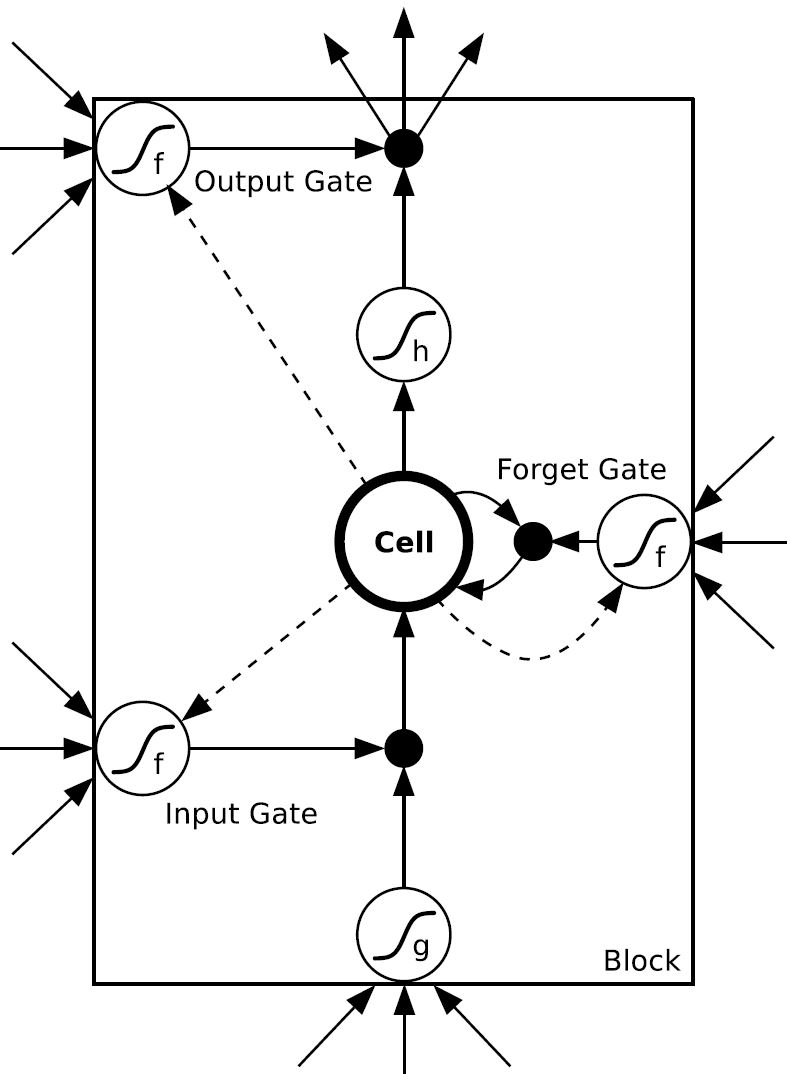}
\caption{A LSTM block (adapted from \cite{lstmD})}
\label{fig:lstmcell}
\end{figure}

To solve the problem of \textquotedblleft vanishing and exploding gradient \textquotedblright, another architecture called Long Short-Term Memory (LSTM) \cite{lstm} was developed. It solves the problem by embedding perfect integrators\cite{pinteg} for memory storage in the network. This is implemented with a complex structure of gates as shown in Fig.~\ref{fig:lstmcell}. For understanding purpose, take a simple perfect integrator $x(t+1) = x(t) + i(t)$, where i(t) is input and x(t) is memory storage. As weight on x(t) here is identity, gradient does not vanish or explode. If we now attach a mechanism to choose when integrator takes the input i.e. a programmable gate depending on context: $x(t+1) = x(t) + g(context)i(t)$, we can now selectively store information for indefinite length of time. Gates in similar sense are used in LSTM to make this possible.           

\section{Approaches} \label{sec:approaches}

\subsection{Neural Turing Machines}
Architecture of Neural Turing Machine (NTM)\cite{ntm} contains mainly: a neural network controller and a memory bank. The controller interacts with outside environment using input vector and output vector. Unlike other neural networks, NTM can also interact with a memory matrix using selective read and write operations (called heads). Every component of this architecture is differentiable, so gradient descent can be applied to train the network.

In NTM an attention mechanism uses \textquoteleft degree of blurriness\textquoteright \ which defines the degree to which the head reads or writes at each memory location. In other words, the head can read or write completely at one location or distributed on many locations. The components of NTM are defined as follows.

\subsubsection{Reading}
$M_t$ is the contents of $N \times M$ memory matrix at time t, where N is the number of memory locations and M the size of memory vector. The model defines $w_t$, a normalized vector over N locations. Normalization of weight vector implies:

\begin{equation}
 \sum_{i=1}^N w_t(i) = 1, \ \ \ \ 0 \leq w_t(i) \leq 1, \forall i
\end{equation}
The length M read vector $r_t$ is defined by following equation:
\begin{equation}
 r_t  =  \sum_{i=1}^N w_t(i)M_t(i)
\end{equation}

\subsubsection{Writing}
Write operation has two parts: an erase followed by an add operation.
For erase operation, the model defines an erase vector $e_t$ whose M elements lie in [0,1]. The old memory $M_{t-1}$ is erased using the following equation:
\begin{equation}
 \tilde{M_t}(i) = M_{t-1}(i)[\textbf{1} - w_t(i)e_t],
\end{equation}
where \textbf{1} is a row-vector of all 1-s, and the multiplication against the memory locations acts point-wise. Memory is erased only when both weighting and erase element at that location are 0.\\
For add operation a length M add vector $a_t$ is defined. It is performed after erase as follows:
\begin{equation}
 M_t(i) = \tilde{M_t}(i) + w_t(i)a_t
\end{equation}
Since both multiplication in erase and addition in add operations are commutative, the order in which multiple heads write is irrelevant. The final memory content $M_t$ is obtained when all heads have done their write operation.
\subsubsection{Addressing Mechanism}
Weightings $w_t$ defined in \textit{Reading}  and \textit{Writing} operations are produced using the addressing mechanism. Two types of addressing mechanism that complement each other are used:
\begin{itemize}
 \item Content-based addressing: focusses attention on memory locations related to values emitted by controller \cite{contentadd}. 
 \item Location-based addressing: For mathematical functions like $f(x,y) = x \times y$, location of the variable is more important than content of the variable. To convey this information location-based addressing is used. 
\end{itemize}
\textbf{Focusing by content}
The model uses a length M key vector $k_t$  produced from head (read or write), a positive key strength $\beta_t$, which can amplify or attenuate the precision of the focus, and a similarity measure $K[.,.]$ (e.g. cosine similarity) between $k_t$ and memory vector $M_t(i)$. These are combined according to following equation to give normalized (using softmax) content based weighting:
\begin{equation}
 w^c_t (i) = \frac{\exp \left( \beta_t K[k_t,M_t(i)] \right)}{\Sigma_j \exp \left( \beta_t K[k_t,M_t(j)] \right)} 
\end{equation}

\textbf{Focusing by Location}
The location-based addressing mechanism facilitates both simple iteration across the memory locations and random access jumps. First, a scalar interpolation gate $g_t$ ($g_t \in [0,1]$) is used to have weighted focus on the content weighting $w_t^c$ and/or the weighting from previous timestep $w_{t-1}$:
\begin{equation}
 w_t^g = g_tw_tĉ + (1-g_t)w_{t-1}
\end{equation}

Second, shift weighting $s_t$ is defined as a normalized distribution over the allowed integer shifts (0 to N-1 memory locations). Then, rotation is applied to $w_t^g$ by $s_t$ by following convolution:
\begin{equation}
 \tilde{w}_t(i) = \sum_{j=0}^{N-1}w_t^g(j)s_t(i-j)
\end{equation}

The combined addressing system can operate in three complementary modes:
\begin{itemize}
 \item weighting chosen only by content system
 \item wighting chosen by content system and then shifted by location system. This allows head to find a contiguous block of data and then, access a particular element within that block. 
 \item weighting from previous timestep is rotated by location system. This allows weighting to iterate through a sequence of addresses by advancing same distance at each timestep.
\end{itemize}

\subsubsection{Controller Network}
Two choices for the neural network to be used as controller network are discussed:
\begin{itemize}
 \item Recurrent controller such as LSTM: Internal memory in this network can be considered as RAM and hidden activations as registers if controller is taken as a CPU. This allows controller to mix information (by unfolding RNN) across multiple time steps of operations.
 \item FeedForward controller: It can mimic recurrent network by reading and writing from the same location in memory at each step. Additionally, these 'read and write operations' on memory matrix are easier to interpret than internal state 'read and write operations' in RNN
\end{itemize}
However, the number of concurrent read and write heads in feedforward controller imposes limitations on type of computation by NTM: with one single read head only unary operations can be performed on memory at each timestep, with two - binary operations, and follows. In RNN, it's taken care of by storing read vectors internally, from previous time steps.

\subsection{Memory Networks}
A memory network\cite{memnn} consists of a memory m and four components:\\
\textbf{I:} input feature map - converts input to internal features\\
\textbf{G:} generalization - updates old memories (state) according to the new input\\
\textbf{O:} output feature map - produces output in feature representation space based on the new input and the current memory state\\
\textbf{R:} response - converts output into desired format
\\
\\
Flow of the model:
\begin{enumerate}
 \item Convert input x to internal input representation I(x)
 \item Update memories m using G
 \item Compute output features o using O
 \item Decode output features o to give the final response
\end{enumerate}

\subsubsection{A MemNN implementation for text}
When neural networks are used as components of a memory network (defined above), it is called memory neural network (MemNN).\\
\textbf{Basic model} Four components of MemNN are defined as follows: \\
\textbf{I:} set of sentences x(question or statement of a fact) transformed as embedding vectors I(x) \\
\textbf{G:} New memories are just stored (no updates) $m_i = I(x)$. Let their number be N memories. \\
\textbf{O:} output features are produced by finding k (taken as 2) supporting memories given x. First memory $o_1$ (k = 1) is retrieved using the following equation:
\begin{equation}
 o_1 = O_1(I(x),m) = arg_{i=1,\cdots,N} s_O(I(x),m_i)
\end{equation}
where $s_O$ is a scoring function on match between I(x) and $m_i$. For k = 2, second memory $o_2$ is found given the first found in previous iteration:
\begin{equation}
 o_2 = O_2(I(x),m) = arg_{i=1,\cdots,N} s_O([I(x),m_{o_1}],m_i)
\end{equation}
where $m_i$ is scored with respect to both the original input and $o_1$, square brackets denote a list. \\
\textbf{R:} It produces a textual response r. Limiting textual response to a single word (out of all words), response is produced by ranking them:
\begin{equation}
 r = argmax_{w \in W} s_R([I(x),m_{o_1},m_{o_2}],w)
\end{equation}
where W is the set of all words in the dictionary, and $s_R$ is a function that scores the match.

\textbf{Training:} It is done in fully supervised setting where desired inputs and responses, and the supporting sentences are labeled as such in the training data (but not in the test data, where only inputs are given). Thus, both $o_1$ and $o_2$ are known at training time. Training is performed with margin ranking loss and stochastic gradient descent (SGD). 

\subsection{End-To-End Memory Networks}
This model\cite{etemn} takes discrete input representations $x_1,\ldots,x_n$ to store them in memory, a query q, and outputs an answer a. Each of the $x_i$, q and \textbf{a} contains symbols from a dictionary with vocabulary V. The model converts x (upto a fixed buffer size) and q to a continuous representation. This representation is processed via multiple hops to output \textbf{a}. As all these representations are continuous we can use backpropagation for training.
\subsubsection*{Single Layer}
In single layer case, the model implements a single memory hop operation.
Structure and flow of single layer model is as follows:\\
\textbf{Input memory representation:} Using embedding matrices A, B (of size d $\times$ V) we convert input x and query q respectively to same continuous space of dimension d. Transformed input is memory vectors \{$m_i$\} and transformed query is u. In the embedding space we compute the similarity between u and $m_i$ by taking the inner product followed by a softmax:
\begin{equation}
 p_i = Softmax(u^T m_i)
\end{equation}
$p_i$ as defined above is probability vector over the inputs.\\
\textbf{Output memory representation:} Using emedding matrix C, each input $x_i$ is transformed to output vector $c_i$. The output response vector o is computed by the following equation:
\begin{equation}
 o = \sum_i p_ic_i
\end{equation}
\textbf{Generating the final prediction:} The predicted label is computed using the final weight matrix W (of size V $\times$ d) by following equation:
\begin{equation}
 \hat{a} = Softmax(W(o + u))
\end{equation}
All three embedding matrices A, B and C, and W are jointly learned during training (Stochastic Gradient Descent) by minimizing a standard cross entropy loss between $\hat{a}$ and the true label a.

\section{Experiment and Results} \label{sec:experiments}

\subsection{Neural Turing Machine} \label{sec:NTM_exp}
Experiments were done on a set of simple algorithms tasks like copying and sorting data sequences. The goal of the experiments was to observe problem solving and learning of compact internal programs by the NTM architecture. Such solution programs could generalize well beyond training data. For example network trained to copy sequences of length 20 was tested on sequences of length 100.
\\
Three architectures are compared in experiments:
\begin{itemize}
 \item NTM with a feedforward controller
 \item NTM with a LSTM controller
 \item Standard LSTM network
\end{itemize}
Tasks were episodic and thus, the dynamic state (previous hidden state) was reset (to learned bias vector) at the start of each input sequence. All the tasks were supervised learning problems; all network had logistic sigmoid output layers and were trained with cross-entropy objective function. Sequence prediction errors are reported in bits-per-sequence.
\subsubsection{Copy}

\begin{figure}
\vspace{-5mm}
\centering
\includegraphics[height=6.2cm]{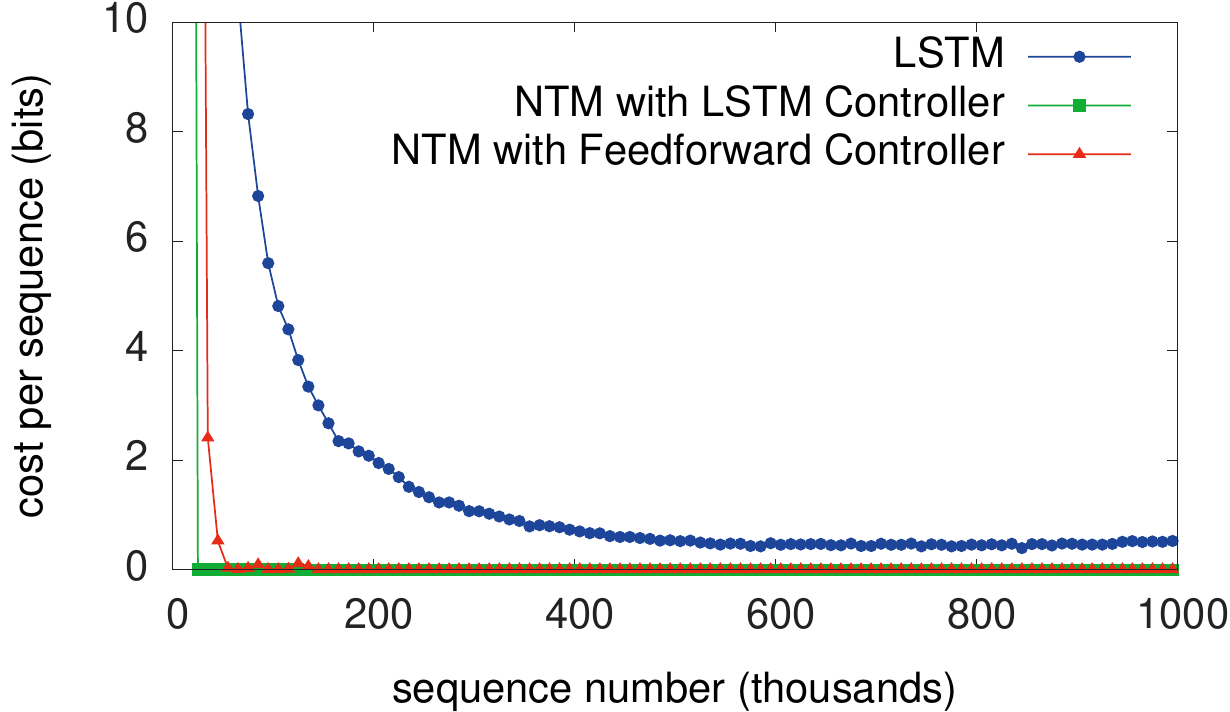}
\caption{Copy Learning Curves (adapted from \cite{ntm})}
\label{fig:ntmcopy}
\vspace{-5mm}
\end{figure}


The task was to copy a sequence of random binary vectors followed by a delimiter (to fix length). Thus, input was a sequence with delimiter and output was the same sequence without delimiter. It was done to compare effect of longer time delays on NTM with LSTM.
\\
As can be seen from Fig. \ref{fig:ntmcopy} NTM learned much faster than LSTM alone, and converged to a lower cost. NTM continues to copy as the length increases while LSTM rapidly degrades beyond length 20. These disparities suggest a qualitative rather than a quantitative difference in problem solving by the two architectures.



\subsubsection{Priority Sort}

\begin{figure}
\vspace{-5mm}
\centering
\includegraphics[width=\columnwidth]{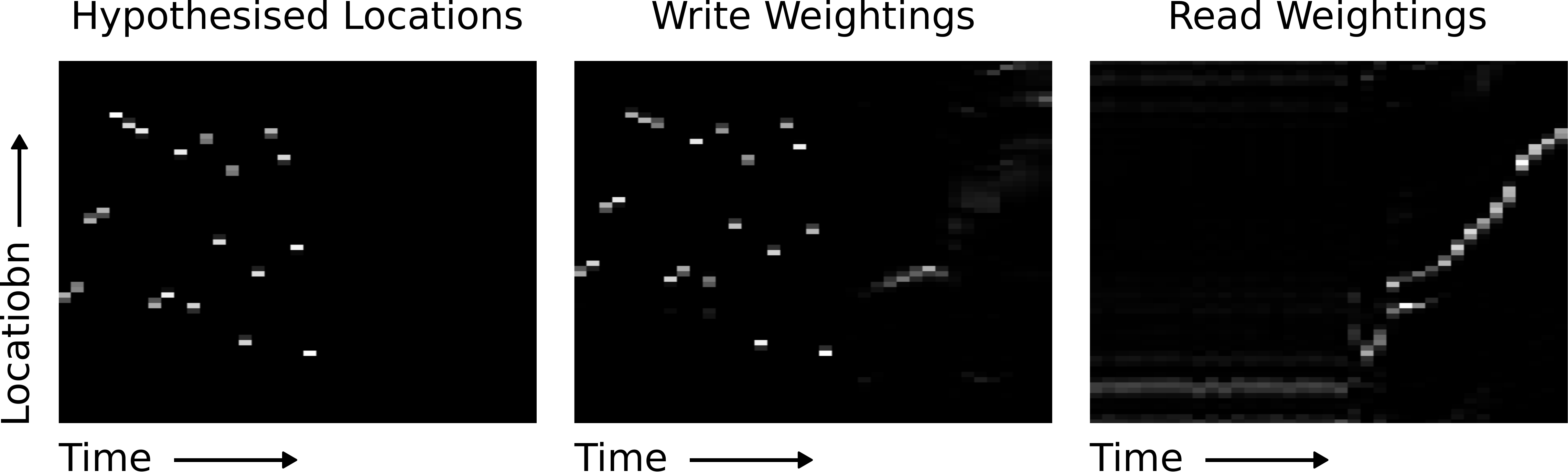}
\caption{ \textbf{NTM Memory Use During the Priority Sort Task.} Left: Write locations returned by fitting a linear function of the priorities to the observed write locations. Middle: Observed write locations. Right: Read locations. (adapted from \cite{ntm})}
\label{fig:ntmsort}
\end{figure}

\begin{figure}
\centering
\includegraphics[height=6.2cm]{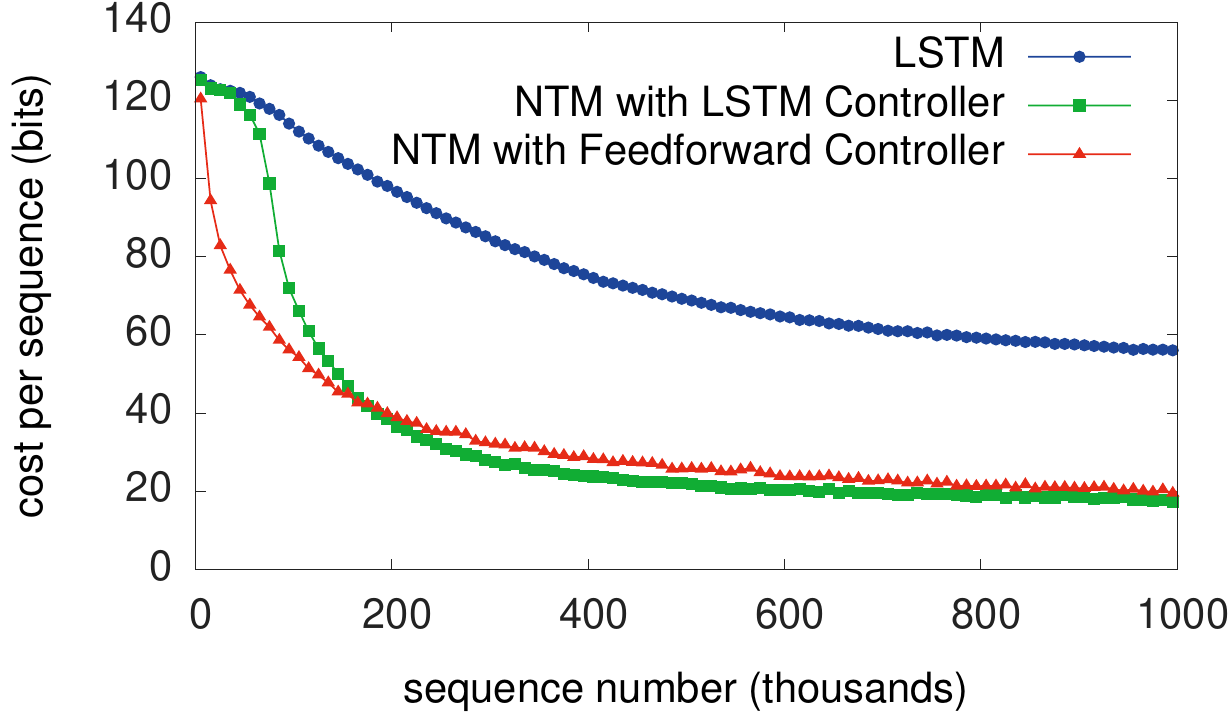}
\caption{Priority Sort Learning Curves. (adapted from \cite{ntm})}
\label{fig:ntmsortcrop}
\vspace{-5mm}
\end{figure}

Sorting capacity of NTM was tested in this task. Input was a collection of random binary vectors with priority from the range [-1,1]. Hypothesis for NTM was that it uses the priorities to determine the relative location of each write. To test the hypothesis a linear function of the priority was fitted on the write locations. Fig. \ref{fig:ntmsort} shows the results, locations returned by the linear function closely match write locations of NTM and reads from the memory locations are in increasing order, thereby sequences were traversed in sorted manner. The learning curves in Fig. \ref{fig:ntmsortcrop} show that NTM outperforms LSTM.   

\subsection{Memory Networks}

\subsubsection{Large-scale QA}

\begin{table}
\vspace{-8mm}
\caption{Results on the large-scale QA task of \cite{qadataset}(adapted from \cite{memnn}).
\label{tab:fader-qa}
}
\begin{center}
\begin{tabular}{|l|c|c|}
\hline
Method & F1 \\
\hline
 Adapted from \cite{qadataset} 
                                         & 0.54 \\
 Adapted from \cite{bordesimpl}            
                                         & 0.73  \\
MemNN (embedding only)       & 0.72  \\
MemNN (with BoW features)    & 0.82 \\
\hline
\end{tabular}
\end{center}
\vspace{-5mm}
\end{table}

Experiments were performed on QA dataset introduced in \cite{qadataset}. It consists of 14M statements, stored as (subject, relation, object) triples. It was combination of pseudo-labeled QA pairs, made of a question and an associated triple, and 35M pairs of paraphrased questions. \\

In experiment framework, top returned candidate answers were re-ranked and results were measured using F1 score over the test set. Systems developed following architecture given in \cite{qadataset} and \cite{bordesimpl}, were tested on the same and compared as shown in Table \ref{tab:fader-qa}. The results show viability of MemNNs in large scale QA answering but the lookup is slow for which method extension like word hashing and cluster hashing were used. 

\subsubsection{Simulated World QA}

\begin{figure}
\vspace{-5mm}
\caption{Example ``story'' statements, questions and answers generated by a simple simulation.
Answering about the location of the milk requires, comprehension of ``picked up'' and ``left'' actions. It also requires comprehension of the time elements of the story, e.g., to answer ``where was Joe before the office?''.
\label{fig:story1}}
\begin{small}
\begin{framed}
Joe went to the kitchen. 
Fred went to the kitchen.
Joe picked up the milk.\\
Joe travelled to the office.
Joe left the milk.
 Joe went to the bathroom.\\
Where is the milk now? \textcolor{red}{A: office}\\
Where is Joe? \textcolor{red}{A: bathroom} \\
Where was Joe before the office? \textcolor{red}{A: kitchen}
\end{framed}
\end{small}
\end{figure}

A simple simulation of 4 characters, 3 objects and 5 rooms - where characters move around, pick up and drop objects, based on the approach of \cite{bordessimul} was built. This simulation was converted into text to form statements and QA. A sample QA is shown in Fig. \ref{fig:story1}. 7K statements and 3K questions were generated for training and the same for testing. MemNNs are compared with RNNs and LSTMs on this task. Difficulty of the task was set based on the limit in the number of time steps (statements), before the entity (in question) was last mentioned. Three kinds of questions were presented to the system separately: about the actor only (actor), about both actor and object, and about actors but without before i.e. not about previous location of actor (actor w/o before). \\

\begin{table}[t]
\vspace{-2mm}
\caption{Test accuracy on the simulation QA task (adapted from \cite{memnn}).
\label{tab:simQA}
}
\begin{center}
\begin{tabular}{|l||c|c|c||c|c|}
\hline
        &    \multicolumn{3}{c||}{Difficulty 1} &  \multicolumn{2}{|c|}{Difficulty 5}  \\
\hline
Method &   actor w/o before & actor & actor+object &  actor & actor+object \\
\hline
RNN                    & 100\% & 60.9\% & 27.9\% & 23.8\% & 17.8\% \\
LSTM                   & 100\% & 64.8\% & 49.1\% & 35.2\% & 29.0\%\\
\hline
MemNN $k=1$           & 97.8\%  &  31.0\% & 24.0\%  &  21.9\% & 18.5\% \\
MemNN $k=1$ {(+time)} & 99.9\%  &  60.2\% & 42.5\%    &  60.8\% & 44.4\% \\
MemNN $k=2$ {(+time)} & 100\%    &  100\% & 100\%  &  100\% & 99.9\% \\
\hline
\end{tabular}
\end{center}
\end{table}

Results for single word answers are given in Table \ref{tab:simQA}. Following observations were made:
\begin{itemize}
 \item RNN and LSTM solved difficulty 1 task w/o before but performed worse \textit{with} before questions and even worse with difficulty 5. This poor performance was attributed to failure in encoding long term memory in RNN, and failure to remember too far sentences in LSTM.
 \item MemNNs did not have above limitations and error was due to wrong statement pick by $s_O$.
 \item Extension of MemNN with time features, based on when a memory slot is written, was essential for such story tasks. 
\end{itemize}

\subsection{End-To-End Memory Networks}
\subsubsection{Synthetic Question and Answering Experiments}
Experiments were performed on synthetic QA tasks defined in \cite{qatasks}. A QA task consists of a set of statements, a question and a corresponding answer. The answers is available at training time and is predicted at testing time. There are 20 different types of tasks that require different forms of reasoning and deduction. Only a subset of provided statements in the task are relevant for answering. This information is not provided to the model at both training and testing time. \\
Following three models (baselines) are compared with this approach (abbreviated MemN2N):
\begin{itemize}
 \item MemNN: The strongly supervised AM+NG+NL Memory Network approach, proposed in \cite{qatasks}. It uses supporting facts (strong supervision), n-gram modelling NG, nonlinear layers NL and an adaptive number of hops AM per query.
 \item MemNN-WSH: A weakly supervised version of MemNN
 \item LSTM: A standard LSTM model, trained using question / answer pairs only (weakly supervised)
\end{itemize}

\begin{table}[h!]
  \begin{adjustwidth}{-15mm}{}
  \vspace{-5mm}
  \centering
  \tiny
  \setlength{\tabcolsep}{5pt}
 \begin{tabular}{|l||c||c|c|c|c|c|c||c|c|c|c|c|}
 \hline
 & \multicolumn{3}{c|}{Baseline} & \multicolumn{9}{c|}{MemN2N} \\ \cline{2-13}
 & Strongly & & & &  & & PE & 1 hop & 2 hops & 3 hops & PE & PE LS \\
 & Supervised & LSTM & MemNN & &  & PE & LS & PE LS & PE LS & PE LS & LS RN & LW \\
Task & MemNN \cite{Weston15} & \cite{Weston15} & WSH & BoW & PE & LS & RN & joint & joint & joint & joint & joint \\
\hline
1: 1 supporting fact        &  0.0 & 50.0 &  0.1 &  0.6 &  0.1 &  0.2 &  0.0 &  0.8 &  0.0 &  0.1 &  0.0 &  0.1 \\
2: 2 supporting facts       &  0.0 & 80.0 & 42.8 & 17.6 & 21.6 & 12.8 &  8.3 & 62.0 & 15.6 & 14.0 & 11.4 & 18.8 \\
3: 3 supporting facts       &  0.0 & 80.0 & 76.4 & 71.0 & 64.2 & 58.8 & 40.3 & 76.9 & 31.6 & 33.1 & 21.9 & 31.7 \\
4: 2 argument relations     &  0.0 & 39.0 & 40.3 & 32.0 &  3.8 & 11.6 &  2.8 & 22.8 &  2.2 &  5.7 & 13.4 & 17.5 \\
5: 3 argument relations     &  2.0 & 30.0 & 16.3 & 18.3 & 14.1 & 15.7 & 13.1 & 11.0 & 13.4 & 14.8 & 14.4 & 12.9 \\
6: yes/no questions         &  0.0 & 52.0 & 51.0 &  8.7 &  7.9 &  8.7 &  7.6 &  7.2 &  2.3 &  3.3 &  2.8 &  2.0 \\
7: counting                 & 15.0 & 51.0 & 36.1 & 23.5 & 21.6 & 20.3 & 17.3 & 15.9 & 25.4 & 17.9 & 18.3 & 10.1 \\
8: lists/sets               &  9.0 & 55.0 & 37.8 & 11.4 & 12.6 & 12.7 & 10.0 & 13.2 & 11.7 & 10.1 &  9.3 &  6.1 \\
9: simple negation          &  0.0 & 36.0 & 35.9 & 21.1 & 23.3 & 17.0 & 13.2 &  5.1 &  2.0 &  3.1 &  1.9 &  1.5 \\
10: indefinite knowledge    &  2.0 & 56.0 & 68.7 & 22.8 & 17.4 & 18.6 & 15.1 & 10.6 &  5.0 &  6.6 &  6.5 &  2.6 \\
11: basic coreference       &  0.0 & 38.0 & 30.0 &  4.1 &  4.3 &  0.0 &  0.9 &  8.4 &  1.2 &  0.9 &  0.3 &  3.3 \\
12: conjunction             &  0.0 & 26.0 & 10.1 &  0.3 &  0.3 &  0.1 &  0.2 &  0.4 &  0.0 &  0.3 &  0.1 &  0.0 \\
13: compound coreference    &  0.0 &  6.0 & 19.7 & 10.5 &  9.9 &  0.3 &  0.4 &  6.3 &  0.2 &  1.4 &  0.2 &  0.5 \\
14: time reasoning          &  1.0 & 73.0 & 18.3 &  1.3 &  1.8 &  2.0 &  1.7 & 36.9 &  8.1 &  8.2 &  6.9 &  2.0 \\
15: basic deduction         &  0.0 & 79.0 & 64.8 & 24.3 &  0.0 &  0.0 &  0.0 & 46.4 &  0.5 &  0.0 &  0.0 &  1.8 \\
16: basic induction         &  0.0 & 77.0 & 50.5 & 52.0 & 52.1 &  1.6 &  1.3 & 47.4 & 51.3 &  3.5 &  2.7 & 51.0 \\
17: positional reasoning    & 35.0 & 49.0 & 50.9 & 45.4 & 50.1 & 49.0 & 51.0 & 44.4 & 41.2 & 44.5 & 40.4 & 42.6 \\
18: size reasoning          &  5.0 & 48.0 & 51.3 & 48.1 & 13.6 & 10.1 & 11.1 &  9.6 & 10.3 &  9.2 &  9.4 &  9.2 \\
19: path finding            & 64.0 & 92.0 &100.0 & 89.7 & 87.4 & 85.6 & 82.8 & 90.7 & 89.9 & 90.2 & 88.0 & 90.6 \\
20: agent's motivation      &  0.0 &  9.0 &  3.6 &  0.1 &  0.0 &  0.0 &  0.0 &  0.0 &  0.1 &  0.0 &  0.0 &  0.2 \\ \hline
Mean error (\%)             &  6.7 & 51.3 & 40.2 & 25.1 & 20.3 & 16.3 & 13.9 & 25.8 & 15.6 & 13.3 & 12.4 & 15.2 \\
Failed tasks (err. $> 5\%$) &    4 &   20 &   18 &   15 &   13 &   12 &   11 &   17 &   11 &   11 &   11 &   10 \\ \hline \hline
On 10k training data        & & & & & & & & & & & & \\
Mean error (\%)             &  3.2 & 36.4 & 39.2 & 15.4 &  9.4 &  7.2 &  6.6 & 24.5 & 10.9 &  7.9 &  7.5 & 11.0 \\
Failed tasks (err. $> 5\%$) &    2 &   16 &   17 &    9 &    6 &    4 &    4 &   16 &    7 &    6 &    6 &    6 \\
\hline
\end{tabular}
\end{adjustwidth}
\vspace{1mm}
  \caption{Test error rates (\%) on the 20 QA tasks for models using
    1k training examples (mean test errors for 10k training examples are shown at the bottom). Key: BoW = bag-of-words representation; PE =
  position encoding representation; LS = linear start training; RN = random injection of time
index noise;  LW = RNN-style layer-wise weight tying (if not stated, adjacent weight tying is used); joint = joint
  training on all tasks (as opposed to per-task training). (adapted from \cite{etemn}) }
\label{tab:MemN2N1}
  
\end{table}

\textbf{Results:} The results across all 20 tasks are given in Table \ref{tab:MemN2N1} for the 1K training set, along with mean performance for 10k training set. Following observations are made:
\begin{itemize}
 \item The best MemN2N models are reasonably close (mean error) to the supervised models.
 \item All variants of MemN2N comfortably beat the weakly supervised baseline methods.
 \item Joint training on all tasks help
 \item More computational hops give improved performance.
\end{itemize}


\section{Conclusion} \label{sec:conclusion}
NTMs enrich the capabilities of recurrent networks most profoundly by using attention mechanism, memory write and a large addressable memory. However, the results of NTMs are only shown on simple tasks of copying and sorting as discussed in section \ref{sec:NTM_exp}. Results of MemNN and MemN2N are compared in Table \ref{tab:MemN2N1}. These suggest, for strong supervision (when supporting facts are known during training) MemNN work the best with least error percentage. But, in case of weak supervision MemN2N are better. It has been consistently observed in all the experiments (Tables \ref{tab:simQA}, \ref{tab:MemN2N1} , Fig. \ref{fig:ntmcopy}, \ref{fig:ntmsortcrop} ) that these new architectures are better in performance than RNN, LSTM for tasks that require large memory lookup for inference. MemN2N have further been applied in many situations like dialogs in a restaurant setting, QA based on a story, goal oriented dialogs etc. These research suggest the prominence of Neural networks in reasoning tasks.

\end{adjustwidth}

\end{document}